\documentclass[letterpaper]{article} 
\usepackage{aaai24}  
\usepackage{times}  
\usepackage{helvet}  
\usepackage{courier}  
\usepackage[hyphens]{url}  
\usepackage{graphicx} 
\usepackage{natbib}  
\usepackage{caption} 
\usepackage{algorithm}
\usepackage{algorithmic}
\usepackage{subfigure}
\usepackage{booktabs}
\usepackage{multirow}
\usepackage{threeparttable}
\usepackage{bm}
\usepackage{amsmath}
\usepackage{amssymb}
\newcommand{\tabincell}[2]{\begin{tabular}{@{}#1@{}}#2\end{tabular}}
\usepackage{enumitem}
\usepackage{newfloat}
\usepackage{listings}
\DeclareCaptionStyle{ruled}{labelfont=normalfont,labelsep=colon,strut=off} 
\floatstyle{ruled}
\newfloat{listing}{tb}{lst}{}
\floatname{listing}{Listing}
\title{Distilling Autoregressive Models to Obtain High-Performance Non-Autoregressive Solvers for Vehicle Routing Problems with Faster Inference Speed}
\author{
   Yubin~Xiao\textsuperscript{\rm 1},
    Di~Wang\textsuperscript{\rm 2,\rm 3}, Boyang~Li\textsuperscript{\rm 4}, Mingzhao Wang\textsuperscript{\rm 1}, \\
    Xuan~Wu\textsuperscript{\rm 1},
    Changliang Zhou\textsuperscript{\rm 5}, 
    You Zhou\textsuperscript{\rm 1}
}
\affiliations{
    \textsuperscript{\rm 1}Key Laboratory of Symbolic Computation and Knowledge Engineering of Ministry of Education, \\College of Computer Science and Technology, Jilin University, China\\
    \textsuperscript{\rm 2}Joint NTU-UBC Research Centre of Excellence in Active Living for the Elderly, \\Nanyang Technological University, Singapore\\
    \textsuperscript{\rm 3}WeBank-NTU Joint Research Institute on Fintech, Nanyang Technological University, Singapore\\
    \textsuperscript{\rm 4}School of Computer Science and Engineering, Nanyang Technological University, Singapore\\
    \textsuperscript{\rm 5}School of System Design and Intelligent Manufacturing, Southern University of Science and Technology, China\\
    xiaoyb21@mails.jlu.edu.cn, wangdi@ntu.edu.sg, boyang.li@ntu.edu.sg, wangmz22@mails.jlu.edu.cn, wuuu20@mails.jlu.edu.cn, 12231198@mail.sustech.edu.cn, zyou@jlu.edu.cn

}

\usepackage{bibentry}

\begin{document}
\hyphenation{stu-dent TSP GNAR-KD  stu-dy mo-de diff-erent  sco-re}

\maketitle

\begin{abstract}
Neural construction models have shown promising performance for Vehicle Routing Problems~(VRPs) by adopting either the Autoregressive~(AR) or Non-Autoregressive~(NAR) learning approach. While AR models produce high-quality solutions, they generally have a high inference latency due to their sequential generation nature. Conversely, NAR models generate solutions in parallel with a low inference latency but generally exhibit inferior performance. In this paper, we propose a generic Guided Non-Autoregressive Knowledge Distillation~(GNARKD) method to obtain high-performance NAR models having a low inference latency. GNARKD removes the constraint of sequential generation in AR models while preserving the learned pivotal components in the network architecture to obtain the corresponding NAR models through knowledge distillation. We evaluate GNARKD by applying it to three widely adopted AR models to obtain NAR VRP solvers for both synthesized and real-world instances. The experimental results demonstrate that GNARKD significantly reduces the inference time~(4$\sim$5$\times$ faster) with acceptable performance drop (2$\sim$3\%). To the best of our knowledge, this study is first-of-its-kind to obtain NAR VRP solvers from AR ones through knowledge distillation.
\end{abstract}

\section{Introduction}
The Vehicle Routing Problem~(VRP) is a well-known optimization problem in the field of transportation and logistics \cite{Kim2015}. VRP has been extensively studied over the past few decades, leading to the development of numerous exact \cite{Applegate2007} and (approximate) heuristics solvers \cite{Helsgaun2017}. Nonetheless, the adoption of VRP solvers in the real world remains challenging when (near) optimal routes need to be determined in real-time with diverse and rapidly changing constraints \cite{Caric2008}. Most conventional VRP solvers are unable to produce high-quality solutions within a reasonably short time period in real-world scenarios \cite{Li2021, Zheng2021}.

As a promising alternative, Neural Networks~(NNs) have been widely applied to solve VRPs in recent years \cite{Hudson2022, Hou2023, Lu2023}. Most NNs used to solve VRPs adopt the Transformer architecture \cite{Vaswani2017} with an encoder-decoder framework, where the decoder produces solutions in an Autoregressive~(AR) manner by generating each node conditioned on the previously generated nodes. Although the state-of-the-art~(SOTA) models \cite{Hou2023} exhibit better performance than conventional VRP solvers. Because the AR process of these SOTA models is not parallelizable and each node is generated using a computationally intensive NN \cite{Joshi2022}, their inference speed becomes much slower as the problem size increases. Furthermore, when employing commonly used sampling techniques like beam search, AR models often suffer from diminishing returns w.r.t. beam size and exhibit limited search parallelism due to the computational dependencies between beams \cite{Koehn2017}.

\begin{figure*}[!t]
\centering
\subfigure[TM]{\includegraphics[width=.68\columnwidth]{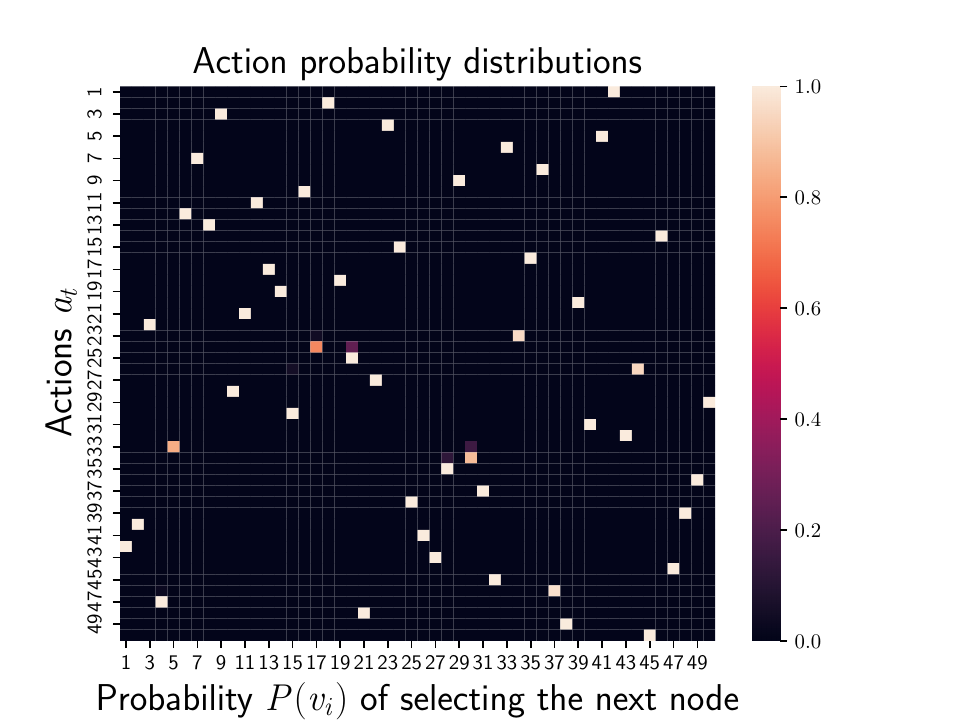}\label{fig5a}}
\subfigure[GCN]{\includegraphics[width=.68\columnwidth]{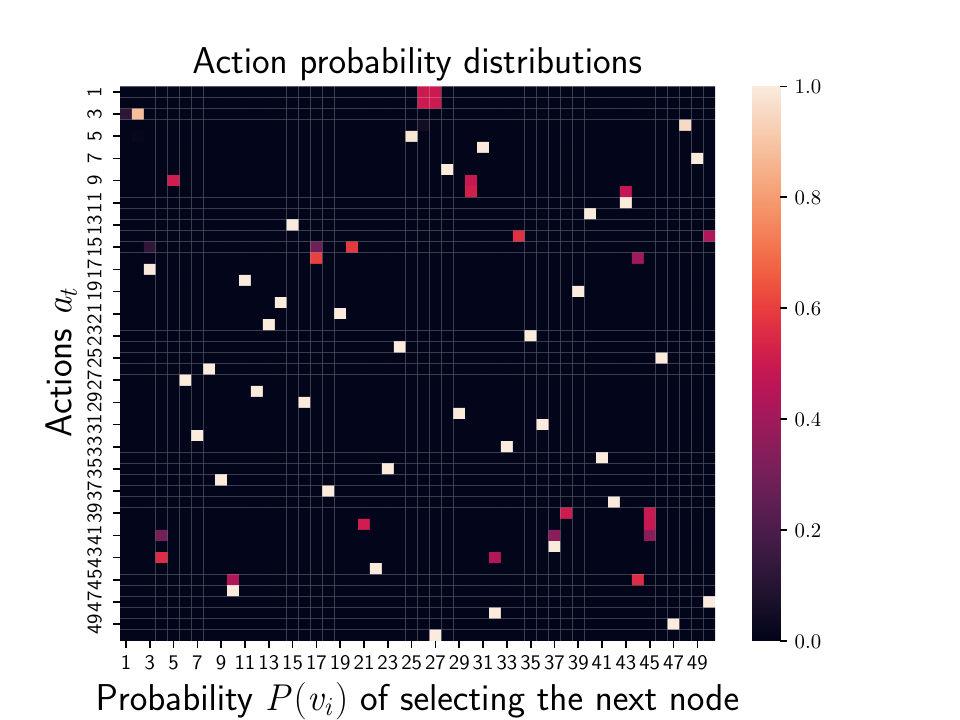}\label{fig5b}}
\subfigure[GNARKD-TM]{\includegraphics[width=.68\columnwidth]{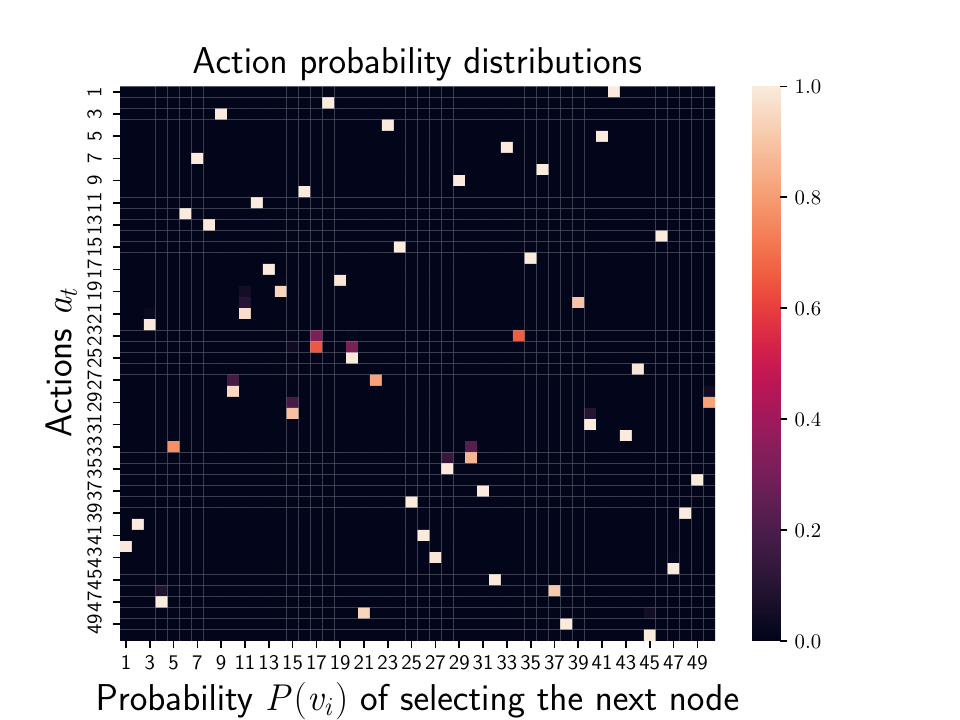}\label{fig5c}}
\caption{Action probability distribution of TM, GCN and GNARKD-TM when solving a randomly generated TSP-50 instance.}\label{fig5}
\end{figure*}

In an effort to improve the inference speed of models for solving VRPs, several Non-Autoregressive~(NAR) models have been developed. These models generate solutions in a one-shot manner, allowing for highly parallelized inference. They treat VRP as a link prediction task and adopt Maximum Likelihood Estimation~(MLE) to maximize the edges' likelihood of being selected in the VRP solution \cite{Joshi2022}. However, most NAR models employ complex Graph Neural Networks~(GNNs) that take both node coordinates and distances between nodes as the inputs, requiring more computationally intensive operations compared to Transformers which only rely on node coordinates. Consequently, the inference speed of NAR models may not meet the anticipated tight timeline. More importantly, NAR models typically exhibit inferior performance compared to AR models. This can be attributed to the lack of order-dependent information during training \cite{Joshi2022} and the NAR model's tendency to make less confident node selections during inference, as reported in our finding (see Figure~\ref{fig5}).

In this paper, we propose a novel method called \textbf{Guided Non-Autoregressive Knowledge Distillation~(GNARKD)} to boost the performance and computational efficiency of NAR models. Inspired by Knowledge Distillation~(KD), which involves transforming knowledge from a complex model (i.e., teacher) to a more compact one (i.e., student). GNARKD converts a Transformer-based AR model into an NAR one using our proposed KD method. Specifically, we make adjustments only to the input/output part of the AR decoder during the KD process, eliminating the necessity for sequential generation of solutions while keeping the AR encoder intact. This approach enables parallel propagation while preserving the pivotal components of the AR architecture. Furthermore, we use the solution generated by the AR model as the teacher network to provide decoding guidance to the NAR model and supervise its decoding output during training. By introducing this training method~(called guided KD), we enable the NAR model to learn information about order dependencies that the conventional NAR models are unable to.

The key contributions of this work are as follows.
\begin{enumerate}[label=(\roman*)]
\item We show that the subpar performance of NAR models in solving VRPs can be attributed to their tendency to take less confident actions during inference. To the best of our knowledge, our work is the first one reporting this finding in the VRP field with the support of experimental results.

\item We propose a novel and generic method named GNARKD to transform AR models into NAR ones to improve the inference speed while preserving essential knowledge. To the best of our knowledge, GNARKD is first-of-its-kind to obtain NAR VRP solvers from AR ones through KD.

\item We apply GNARKD to three widely adopted AR models and evaluate their performance using the widely used synthesized and real-world datasets. The experimental results demonstrate that GNARKD significantly reduces the inference time and achieves on-par solution quality comparing with the teacher models. This finding suggests that the derived NAR models are suitable for deployment in real-world scenarios that demand instantaneous, near optimal solutions.
\end{enumerate}

\section{Related Work}
With the emergence of deep learning, numerous NN-based model have been proposed in recent years to solve VRPs \cite{Fu2021, Zong2022, Zhang2022}. These models can be generally categorized into neural improvement type and neural construction type. We focus on the latter in this paper because the former typically has longer inference latency~(e.g., \cite{Wu2022}), which is contrary to our strive for shorter inference time.

Neural construction VRP solvers can be classified as either AR (majority) or NAR (minority) based on the construction methods. AR models produce VRP solutions incrementally by generating one node at each decoding state. For instance, \citeauthor{Kool2019} (\citeyear{Kool2019}) proposed the widely recognized Attention Model~(AM) that employs Transformers to solve VRPs. \citeauthor{Kwon2020} (\citeyear{Kwon2020}) extended AM by introducing the Policy Optimization with Multiple Optima~(POMO) model and yielded SOTA results.

While AR models yield high-performance results~\cite{Jin2023, Pan2023}, their sequential approach to generating solutions typically results in slower inference speed compared to NAR models \cite{Joshi2022}, primarily because NAR models generate solutions in a one-shot manner \cite{Wang2023}. For instance, \citeauthor{Joshi2019} (\citeyear{Joshi2019}) utilized Supervised Learning (SL) to train a GNN-based model, handling TSP as a link prediction problem, and employed greedy and beam search algorithms to generate solutions. However, this NAR model, despite its high inference speed, neglects the importance of the sequential ordering in TSP tours, resulting in suboptimal performance. \citeauthor{Xiao2023} (\citeyear{Xiao2023}) introduced the first Reinforcement Learning~(RL)-based NAR model for solving TSPs and exhibited satisfactory performance. However, this proposed architecture is only applicable to TSP. Moreover, most NAR models employ GNNs that consider both node coordinates and distance matrices as the model inputs, as opposed to Transformers that solely rely on node coordinates. The adoption of GNN could potentially hinder the acceleration of inference speed, making it challenging for these models to meet an anticipated tight timeline. Currently, there is a lack of Transformer-based NAR model designed to tackle VRPs.

AR and NAR models follow distinct research approaches, without a well-defined mechanism to leverage their respective advantages~(i.e., high solution quality of the former and low inference latency of the latter). Therefore, we advocate for a generic method that leverages the strengths of existing Transformer-based AR models to improve the performance of NAR models in solving VRPs.

\section{Preliminaries and Study Motivation}
This section presents the formulation of VRPs and introduces the motivation of this study.


\subsection{VRP Setting}
\noindent We define a VRP-$n$ instance as a graph comprising $n$ nodes in the node set $V$. Each node is denoted as $v_i\in V$. The optimal solution of a VRP is the tour $\pi^*$ that traverses through all nodes with the shortest overall length. Solving different VRP variants may be subject to various problem-specific constraints. This study specifically examines two prominent VRP variants:~TSP and Capacitated VRP~(CVRP), due to their representativeness and widespread applications in various domains \cite{Kim2015}. In TSP, a feasible tour entails visiting each node in $ V$ exactly once. CVRP extends TSP by introducing an additional depot node $v_0$, a capacity constraint, and demand requests of each node that are smaller than the capacity constraint. A feasible tour for CVRP consists of multiple sub-tours that each visits a subset of nodes and serves the demand not exceeding the capacity. All nodes, except for the depot, must be visited exactly once.

\subsection{Tour Generation by AR and NAR Models}\label{motivation}
\subsubsection{AR Models.} These models are commonly employed to solve VRPs using a Markov Decision Process~(MDP) and utilize a Transformer-based encoder-decoder framework as its policy network. The encoder captures node features and the decoder generates a tour~$\pi$ based on the extracted features and the action history of node selections $a_{1:t-1}$. At each step $t$ of the MDP, the decoder takes an action $a_t$ to choose an unvisited node, masking invalid nodes~(visited and exceeding capacity) to ensure feasibility, until the tour is completed. Given a VRP instance $s$, the AR process can be factorized into a chain of conditional probabilities as follows:
\begin{equation}\label{eq1}
P_{\textit{AR}}(\pi^{\textit{AR}}|s,\theta^{\textit{AR}})=\prod_{t=1}^{l}p_{\textit{AR}}(a_t|a_{1:t-1},s,\theta^{\textit{AR}}),
\end{equation}
where $p_{\textit{AR}}$ denotes the policy parameterized by $\theta^{\textit{AR}}$ and $l$ denotes the number of actions taken to complete the tour. For TSP, $l=n$, while $l\geq n$ for CVRP because the depot is visited at least once.

\subsubsection{Pros and Cons of AR Models.} The AR process simulates the locomotion and observation involved in human decision-making, and can produce near-optimal VRP solutions. However, the individual steps of the AR decoder must be executed sequentially rather than in parallel, which undermines the advantage of the Transformer in terms of inference speed \cite{Gu2018}. Moreover, the sequential nature of AR models leads to restricted search parallelism and diminishing returns w.r.t. beam size in beam search \cite{Koehn2017}.

\subsubsection{NAR Models.} These models eliminate the sequential dependencies between nodes in the tour, resulting in faster inference speed and robust parallelism in the inference process. NAR models usually adopt deep learning to generate a score matrix, which quantifies the likelihood of each edge being selected in the tour, and gradually take an action of choosing an edge to construct a feasible tour. Notably, although almost all existing NAR VRP models are designed to solve TSPs, they are easily extensible to solve CVRPs with a trivial modification in the inference method~(see Section~\ref{Feasibility}). The NAR process can be factorized into a product of independent probabilities as follows:
\begin{equation}\label{eq_nar}
P_{\textit{NAR}}(\pi^{\textit{NAR}}|s,\theta^{\textit{NAR}})=\prod_{t=1}^{l}p_{\textit{NAR}}(a_t|s,\theta^{\textit{NAR}}).
\end{equation}

\begin{figure*}[!t]
	\centering
	\includegraphics[width=2\columnwidth]{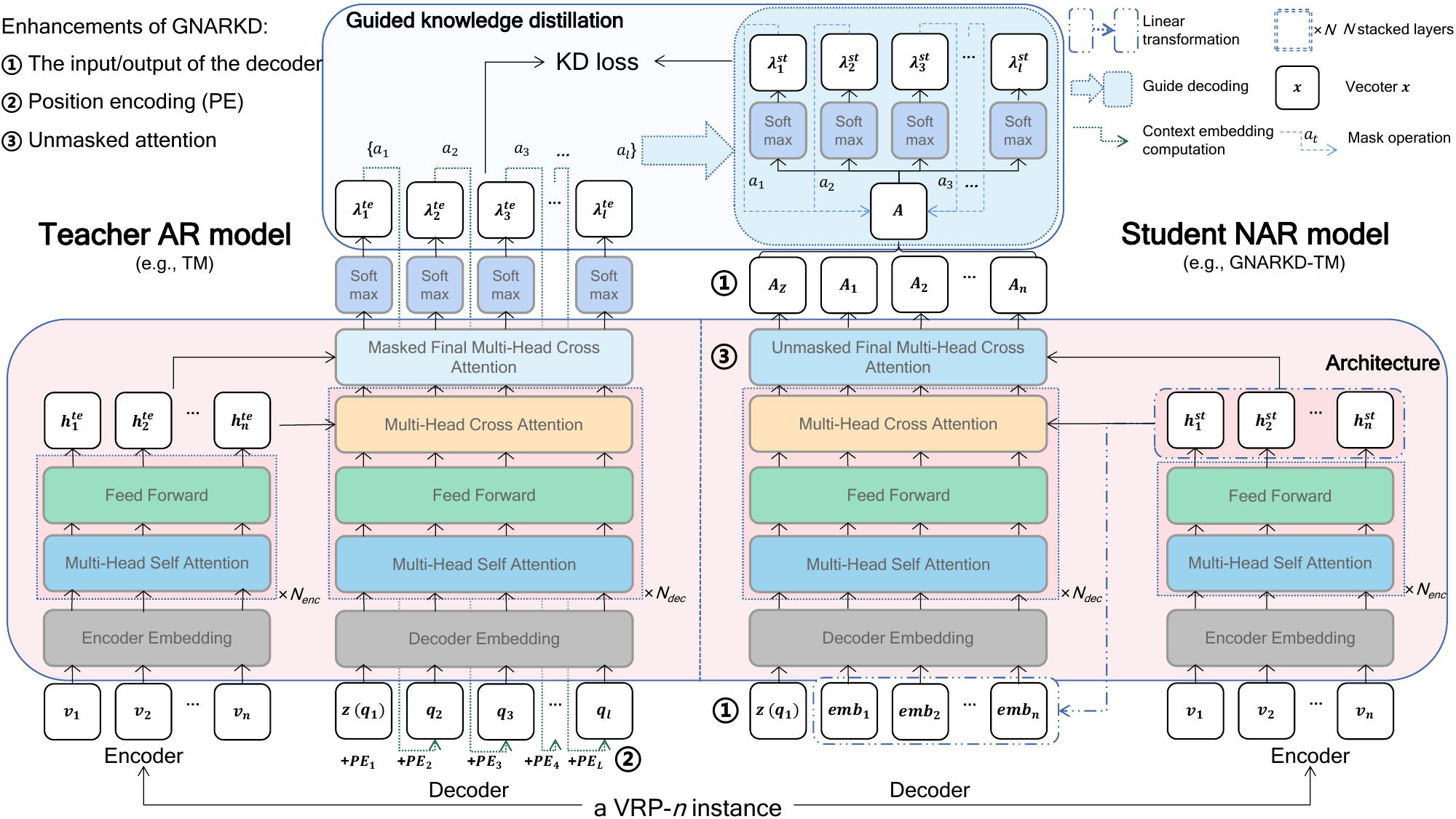}
	
	\caption{The architecture of GNARKD, which transforms the teacher AR model into a student NAR model.}\label{main_fig}
\end{figure*}
\subsection{Challenge and Motivation}
\noindent The performance of NAR models in terms of the overall tour length is usually not satisfactory. To investigate this phenomenon, we analyze the probability distributions of actions in both AR and NAR models. Specifically, we select the pre-trained TM \cite{Bresson2021} and GCN \cite{Joshi2019} models as representatives of AR and NAR models, respectively. We visualize the probability distribution of selecting the next node in solving a randomly generated TSP-50 instance for both models, using the same inference method of greedy search. Figures~\ref{fig5a} and~\ref{fig5b} show that TM produces a highly deterministic solution (majority data points are light-colored), while GCN achieves partial determinism (minority data points are light-colored). Our statistical analysis indicates that TM's actions have a lower bound probability of 0.75, while GCN's actions have a much smaller lower bound probability of 0.49. Furthermore, when using a probability threshold of 0.75 to identify confident actions, only 76\% of the actions in GCN meet the criterion, indicating that the NAR model generates less confident actions during the construction of tours. Because the primary objective of training neural construction models is to increase the probability of generating high-quality solutions \cite{Hottung2022}, the suboptimal quality of solutions in NAR model can be naturally attributed to the diminished probability while generating these solutions. Based on these findings, we propose to leverage the confident actions of the AR model to enhance the NAR model's performance while preserving the fast inference speed.

\section{Guided NAR Knowledge Distillation}
To combine the advantages of both AR and NAR models, we introduce GNARKD, which transforms a Transformer-based AR model into an NAR one through KD.

We show the architecture of GNARKD in Figure~\ref{main_fig}. For a given pre-trained Transformer-based AR model~(e.g., TM), we modify the decoder's input/output part, eliminating its constraint of sequential generation (i.e., node selection no longer depends on the action history). This modification leads to the development of an NAR model, the student in our KD approach. Subsequently, we conduct guided KD, allowing the student to replicate actions generated by the teacher, while aligning the student's action probability distribution with that of the teacher for training.

\subsection{Architecture of the Student Model}

\noindent To construct the student model, we convert the network architecture of a Transformer-based AR model into its NAR form. Without the loss of generality, in the following subsections, we describe our enhancements to a vanilla Transformer architecture that used for most existing AR models. This architecture includes an encoder and a decoder comprising only feed-forward layers and multi-head attention modules. Thus, the absence of recursive structures such as RNN eliminates the inherent necessity for sequential execution, thereby enabling the development of NAR models.

\subsubsection{Student Encoder Stack.} To facilitate the transformation of knowledge from the teacher to the student, we refrain GNARKD from modifying the encoder, thus promoting a proximity of representations for identical nodes in both models. Formally, we define the node representation by the teacher and the student as $\bm{h^{\textit{\textbf{te}}}}\in\mathbb{R}^{d_h\times n} $ and $\bm{h^{\textit{\textbf{st}}}}\in\mathbb{R}^{d_h\times n}$, respectively, both with $d_h$ hidden dimensions.

\subsubsection{Student Decoder Stack.} All three key enhancements are centered around the decoder as described as follows.

\noindent\textbf{1)~The input and output of the decoder.} The decoder in AR models is responsible for generating the action probability distribution~$\bm{\lambda^{\textit{\textbf{te}}}_t}$ that aids in selecting the subsequent node at time $t$, based on the node representation~$\bm{h^{\textit{\textbf{te}}}}$ and the action history of previously selected nodes~$a_{1:t-1}$. This process is generally described as follows:
\begin{equation}
\bm{\lambda^{\textit{\textbf{te}}}_t}=\sigma_{T_1}\left( \operatorname{f^{\text{te}}}\left(\begin{cases}
		 \bm{z}, & \text{ if } t = 1, \\
		\operatorname{CE}(a_{1:t-1}), & \text{ if } t > 1,
	\end{cases}, \bm{h^{\textit{\textbf{te}}}}\right)\right),
\end{equation}
where $\bm{\lambda^{\textit{\textbf{te}}}_t}\in\mathbb{R}^{1\times n}$, $\sigma_{T_1}$ denotes the Softmax function with temperature $T_1$, $\operatorname{f^{\text{te}}}$ denotes the forward propagation function of the teacher's decoder, $\bm{z}\in\mathbb{R}^{d_h\times 1}$ denotes a learnable parameter acting as the input placeholder~(or the depot in CVRP), and $\operatorname{CE}(a_{1:t-1})$ denotes the context embedding based on the action history of node selections~$a_{1:t-1}$. Different AR models may compute the context embedding using different methods with the same purpose of constructing a query $\bm{q_t}\in\mathbb{R}^{d_h\times 1}$ for selecting node at time $t$.

To obtain the NAR form of the student decoder that enables parallel propagation, we omit the inclusion of the action history information and instead model the correlation between each node and its immediate successor. Furthermore, we enhance the versatility of GNARKD by adopting a consistent and simple query construction method across different GNARKD student models. Specifically, we apply a shared linear transformation parameterized by weights $\bm{W}$ to each node embedding $\bm{h^{\textit{\textbf{st}}}_i}$, denoted as $\bm{\textit{\textbf{emb}}_i}=\bm{W}\bm{h^{\textit{\textbf{st}}}_i}, \bm{\textit{\textbf{emb}}_i}\in\mathbb{R}^{d_h\times 1}$, which functions as the query for selecting node at time $t$. Subsequently, the decoder computes the tightness $A_{i,j}$ of one node $v_i$ connected to another node $v_j$ through the forward propagation function $\operatorname{f^{\text{st}}}$ of the student \cite{Joshi2019}. This parallel processing is achieved by concatenating all node representations as follows:
\begin{equation}
\bm{A} = \operatorname{f^{\text{st}}}(\bm{W}[\bm{z},\bm{h^{\textit{\textbf{st}}}_1},\dots,\bm{h^{\textit{\textbf{st}}}_n}]), \bm{A}\in\mathbb{R}^{(n+1)\times n},
\end{equation}
where $[\cdot ,\cdot,\cdot]$ denotes the horizontal concatenation operator.

\noindent\textbf{2)~Position encoding.} Because the student decoder only considers individual nodes and not their positions in the tour~(at least within the decoder), we refrain the student model from using positional encoding that is commonly used in Transformer-based AR decoders.

\noindent\textbf{3)~Unmasked attention.} Because the AR model's output is constrained by sequential dependencies, its decoder relies on masks within the final cross attention layer to avoid revisiting selected nodes. Conversely, the student decoder does not employ such masks because it only relies on the post-decoder output processing for generating valid solutions.

\subsection{Guided Knowledge Distillation}\label{training}


\noindent To effectively transform the knowledge from a AR model to an NAR one without losing the order dependence information, we introduce guided KD for the NAR student to preserve the inherent order dependence information of the solutions generated by AR models. Specifically, we employ the teacher's actions to guide the decoding process of the student's output. Thus, the NAR student models are trained by learning a proxy distribution derived from the action probability distribution of the AR teacher models, effectively preserving the order dependencies.


\subsubsection{Guided Decoding.} We require the student to replicate the identical sequence of actions $\pi^{\textit{AR}}$ generated by the teacher. Then, for each action $a_t\in \pi^{\textit{AR}}$, we apply the Softmax function $\sigma_{T_2}$ with temperature $T_2$ to normalize the tightness $A_{a_t, i}$ of the connection between the selected node of that action and the other node $v_i$, thereby obtain the probability distribution of actions for the student at time $t$ as follows:
\begin{equation}
\bm{\lambda_t^{\textit{\textbf{st}}}}=\sigma_{T_2}\left (A_{a_t, i}\odot \begin{cases}
		-\infty, & \text{ if node } v_i  \in \textit{set}_t, \\
		1, & \text{ otherwise},
	\end{cases}  \right ),
\end{equation}
where $i\in \{1,\dots,n\}$, $\bm{\lambda_t^{\textit{\textbf{st}}}}\in\mathbb{R}^{1\times n}$, $\odot$ denotes the element-wise multiplication, and $\textit{set}_t$ denotes the set of constrained nodes at time $t$ (see Section~\ref{Feasibility} for details).

\subsubsection{Learning the Proxy Distribution.}
We use the teacher's action probability distribution $\bm{\lambda^{\textit{\textbf{te}}}_t}$ as the supervision signals for training the student. To promote the student's confidence in taking actions by learning a leptokurtic probability distribution \cite{Zhang2023}, we use the Softmax function with a temperature $T_1\textless 1$ for the teacher model, while setting the student model's temperature to $T_2=1$. Formally, we optimize the student model's learnable parameters $\theta$ by minimizing the KL divergence between its action probability distribution and that of the teacher as follows:
\begin{equation}
\begin{aligned}
\mathcal{L}_{\textit{KD}}&=\mathbb{E}_{\pi^{\textit{AR}}\sim P_{\textit{AR}}(\cdot|s)}[\operatorname{KL}[{P_{\textit{AR}}(\pi^{\textit{AR}}|s)}||{P_{\textit{NAR}}(\pi^{\textit{AR}}|s, \theta)}]]\\
&= \mathbb{E}_{\pi^{\textit{AR}}\sim P_{\textit{AR}}(\cdot|s)}\left[\operatorname{KL}\left[\prod_{t=1}^{l}\bm{\lambda^{\textit{\textbf{te}}}_t}||\prod_{t=1}^{l}\bm{\lambda_t^{\textit{\textbf{st}}}}\right]\right]        \\
&=\frac{1}{B}\sum^{B} \sum_{t=1}^{l} \sum_{i=1}^{n} \lambda_{t,i}^{\textit{te}} \operatorname{log}(\lambda_{t,i}^{\textit{te}}-\lambda_{t,i}^{\textit{st}}),
\end{aligned}
\end{equation}
where $B$ denotes the batch size used during training, $\lambda^{\textit{te}}_{t, i}$ and $\lambda_{t,i}^{\textit{st}}$ denote the action probability of selecting the node $v_i$ at time $t$ for the teacher and student, respectively. The source code of GNARKD is accessible online\footnote{https://github.com/xybFight/GNARKD}.

\subsection{Solving CVRPs in an NAR Manner}\label{Feasibility}
To date, only a handful of NAR models have been proposed to address VRPs \cite{Joshi2019, Xiao2023}, all of which focus exclusively on solving TSPs. We introduce the search method that uses an NAR model to solve CVRPs. Specifically, given the matrix $A$ output by the NAR model, we generate the VRP solutions via greedy search at step $t$ as follows:
\begin{equation}
a_t = \begin{cases}
        v_{\textit{st}}, & \mbox{if } t=1,  \\
        \operatorname{argmax}(\bm{\lambda_{a_{t-1}}}), & \mbox{otherwise}.
      \end{cases}
\end{equation}
\begin{equation}
\bm{\lambda_{a_t}} =\sigma_{T_2}\left (A_{a_t,i}\odot \begin{cases}
		-\infty, & \text{ if node } v_i  \in set(t), \\
		1, & \text{ otherwise},
	\end{cases}\right ),
\end{equation}
where $i\in \{1,\dots,n\}$, $\bm{\lambda_{a_t}}\in\mathbb{R}^{1\times n}$, and $v_{\textit{st}}$ denotes the starting point of the VRP solution. For TSPs, we add all the visited nodes to $set(t)$:
\begin{equation}
set(t)=\{a_1, a_2, \dots, a_{t-1}\}.
\end{equation}
For CVRP, we augment $set(t)$ to incorporate nodes whose demand exceeds the remaining vehicle capacity, so as to satisfy the relevant constraints:
\begin{equation}
set(t)=\{a_1, a_2, \dots, a_{t-1}\}\cup \{\begin{cases}
		a_i, & \text{ if }  \delta_{i}\textgreater Q_t  , \\
		\emptyset , & \text{ otherwise},
	\end{cases}\},
\end{equation}
where $\delta_{i}$ denotes the demand of node $v_i$ and $Q_t$ represents the remaining vehicle capacity at time $t$. At time $t=1$, $Q_1$ is initialized as $Q_1 = Q$, after which it is updated as follows:
\begin{equation}
Q_{t+1}={\begin{cases}
       Q, & \mbox{if } v_{a_t}=v_0, \\
       Q_t-\delta_{i}, & \mbox{otherwise}.
     \end{cases}}
\end{equation}

\begin{table*}[!t]
  \centering
    \resizebox{2\columnwidth}{!}{
    \begin{tabular}{cc|c|ccc|ccc}
    \toprule
      & \multirow{3}{*}{\textbf{Teacher and student}} & \multirow{3}{*}{\textbf{Inference mode}} & \multicolumn{3}{c}{$\bm{n=50}$} & \multicolumn{3}{c}{$\bm{n=100}$} \\
      & &  & \textbf{\tabincell{c}{Learning\\gap}}$\downarrow$ & \textbf{\tabincell{c}{Acceleration\\ratio (S)}}$\uparrow$ & \textbf{\tabincell{c}{Acceleration\\ratio (T)}}$\uparrow$ & \textbf{\tabincell{c}{Learning\\gap}}$\downarrow$ & \textbf{\tabincell{c}{Acceleration\\ratio (S)}}$\uparrow$ & \textbf{\tabincell{c}{Acceleration\\ratio (T)}}$\uparrow$ \\
    \cmidrule{1-9}
    \multirow{8}[6]{*}{\rotatebox{90}{TSP}} & \multirow{3}[0]{*}{AM and GNARKD-AM} & Greedy & 1.90\% & 6.58$\times$ & 11.7$\times$ & 3.06\% & 7.33$\times$ & 22.47$\times$ \\
    &  & Sampling (1000) & -0.13\% & 3.59$\times$ & 101.70$\times$ & 0.18\% & 4.54$\times$ & 69.39$\times$ \\
    &  & Sampling (2000) & -0.15\% & 3.51$\times$ & 76.80$\times$ & 0.14\% & 5.95$\times$ & 57.04$\times$ \\
    \cmidrule{2-9}
    & \multirow{2}[0]{*}{POMO and GNARKD-POMO} & Greedy (no augment) & 0.26\% & 3.62$\times$ & 2.93$\times$ & 0.69\% & 3.77$\times$ & 4.50$\times$ \\
    &  & Greedy ($\times$ 8 augment) & 0.01\% & 3.31$\times$ & 3.47$\times$ & 0.11\% & 3.35$\times$ & 5.23$\times$ \\
        \cmidrule{2-9}
        & \multirow{3}[0]{*}{TM and GNARKD-TM} & Greedy & 1.06\% & 9.12$\times$ & 4.38$\times$ & 6.34\% & 9.89$\times$ & 9.57$\times$ \\
    &  & Sampling (1000) & -0.02\% & 6.83$\times$ & 156.86$\times$ & 0.00\% & 11.25$\times$ & 179.40$\times$ \\
    &  & Sampling (2000) & -0.02\% & 9.81$\times$ & 159.51$\times$ & -0.02\% & 17.43$\times$ & 170.18$\times$ \\
    \midrule
    \multirow{5}[6]{*}{\rotatebox{90}{CVRP}} &\multirow{3}[0]{*}{AM and GNARKD-AM} & Greedy & 10.28\% & 3.12$\times$ & 7.03$\times$ & 12.03\% & 3.15$\times$ & 5.88$\times$ \\
    &  & Sampling (1000) & 1.68\% & 1.83$\times$ & 97.87$\times$ & 4.26\% & 2.55$\times$ & 71.44$\times$ \\
    &  & Sampling (2000) & 1.43\% & 2.13$\times$ & 56.36$\times$ & 3.95\% & 3.54$\times$ & 52.62$\times$ \\
     \cmidrule{2-9}
     & \multirow{2}[0]{*}{POMO and GNARKD-POMO} & Greedy (no augment) & 3.02\% & 2.04$\times$ & 3.52$\times$ & 7.10\% & 2.27$\times$ & 4.97$\times$ \\
    &  & Greedy ($\times$ 8 augment) & 1.73\% & 2.06$\times$ & 3.92$\times$ & 4.51\% & 3.35$\times$ & 4.96$\times$ \\
     \midrule
   \multicolumn{3}{c|}{\textbf{Average}}& \textbf{1.61\%}& $\bm{4.42\times}$& $\bm{52.77\times}$ &\textbf{3.25\%} & $\bm{6.03\times}$ & $\bm{50.59\times}$ \\
    \bottomrule
    \end{tabular}}
    \caption{Performance of GNARKD on solving VRPs with comparisons. The learning gap is obtained by taking the difference between the solution quality of the student and that of the teacher with reference to the teacher; the acceleration ratio is obtained by the student relative to the teacher in terms of solving time for a single instance (S) and for total 10,000 instances (T). Because TM lacks a dedicated method for solving CVRPs, we exclude all the corresponding comparisons regarding CVRP in this paper. Moreover, due to the GPU memory limitation (8GB) and the ability to simultaneously process multiple input instances of each model, we run them with the largest batch size possible for all the comparisons in this paper when solving all 10,000 instances.}
  \label{table1}%
\end{table*}%

\section{Experimental Results}\label{experiment}
We comprehensively evaluate the performance of GNARKD on TSPs and CVRPs, using a computer equipped with an Intel(R) i5-11400F CPU and an NVIDIA RTX 3060Ti GPU.

\subsection{Experiment Setup}
\subsubsection{Synthetic Data Generation.} For TSPs, we randomly sample node coordinates from a uniform distribution on the unit interval, as done in prior studies \cite{Kwon2020}. For CVPRs, we follow the instance generation process in \cite{Kwon2020}. Specifically, we sample the coordinates of each node and the depot uniformly at random from the unit interval. The demand $\delta_{i}$ for each node $v_i$ is defined as $\delta_{i} = \frac{\hat{\delta_{i}}}{\lfloor \frac{n}{5}\rfloor +30}$, where $\hat{\delta_{i}}$ is a uniformly sampled discrete value from \{1,$\dots$,9\}. The capacity $Q$ is set to 1.

\subsubsection{Teacher Model.} We apply GNARKD to three prominent AR models, namely AM \cite{Kool2019}, POMO \cite{Kwon2020} and TM \cite{Bresson2021}, and name the obtained NAR models GNARKD-\{AM, POMO, TM\}, respectively.

\subsubsection{Student Model.} To make a fair comparison, we maintain the consistency in the model size and hyperparameter values for each student and its respective teacher. Furthermore, we initialize the weights of the student's encoder with those of its teacher because both models share the same encoder input and architecture.

\subsubsection{Training and Hyperparameters.} We maintain a consistent distillation environment for all student models to simplify and standardize training conditions. Specifically, we set the temperature $T_1$ of teacher model to $0.1$ based on preliminary experimental results. Each training epoch consists of $1000$ batches of $100$ instances generated randomly on the fly. The total number of training epochs varies according to the problem size, with $500$ and $1000$ epochs used for $n=50$ and $n=100$, respectively. We use the outputs of the teacher in ``greedy mode" as the supervisory signals for its student. We utilize Adam \cite{Kingma2014} optimizer with a fixed learning rate of 1e--5.

\subsubsection{Inference.} We adopt the two standard inference modes, namely ``greedy" and ``sampling", for both the student and teacher models. Specifically, the ``greedy mode" employs the $\operatorname{argmax}$ function to generate a single deterministic solution. This process starts at the depot or a designed placeholder $\bm{z}$. The next node $v_{a_{t+1}}$ is selected greedily based on the largest tightness $\bm{A_{a_t}}$ among all the valid neighboring edges. The search ends when all nodes have been visited. On the other hand, the ``sampling mode" generates multiple solutions by following a probabilistic policy. One well-known sampling method is beam search, which is a breadth-first search with width $B$. Beam search commences its exploration from initial candidates (e.g., the depot node $v_0$) and expands the tour by evaluating $B$ potential successors. At each stage (time $t$), the top-$B$ sub-tours with the highest probability $\prod_{1}^{t}p(a_t)$ are iteratively updated and retained until all nodes have been visited. Besides, in our comparison of GNARKD-POMO and its teacher model, we employ the greedy multiple rollouts technique with and without $\times8$ augments as used in POMO.

\begin{table*}[!t]
  \centering
    \resizebox{2\columnwidth}{!}{
    \begin{tabular}{cc|c|c|ccc|ccc}
    \toprule
    & \multirow{3}{*}{\textbf{Method}} & \multirow{3}{*}{\textbf{Type}} & \multirow{3}{*}{\textbf{Inference mode}} & \multicolumn{3}{c}{$\bm{n=50}$}  & \multicolumn{3}{c}{$\bm{n=100}$}  \\
      & &   &  & \textbf{\tabincell{c}{Average\\length}}$\downarrow$ & \textbf{\tabincell{c}{Optimality\\gap}}$\downarrow$ & \textbf{\tabincell{c}{Time$^*$\\(sec)}}$\downarrow$  & \textbf{\tabincell{c}{Average\\length}}$\downarrow$ & \textbf{\tabincell{c}{Optimality\\gap}}$\downarrow$ & \textbf{\tabincell{c}{Time$^*$\\(sec)}}$\downarrow$ \\
    \midrule
    \multirow{10}[6]{*}{\rotatebox{90}{TSP}}   & Concorde & exact solver & - & \textbf{5.689} & \textbf{0.00\%} & \textbf{0.035}, ($3.63\times10^2$) & \textbf{7.765} & \textbf{0.00\%} & 0.165, ($1.45\times10^3$) \\
    & MDAM & AR & Sampling (30) & 5.690 & 0.03\% & 18.52, ($4.74\times10^2$) & 7.798 & 0.42\% & 32.46, ($1.56\times10^3$) \\
    & GCN & NAR & Sampling (1280) & 5.710 & 0.37\% & 0.073, ($1.45\times10^2$) & 7.920 & 2.00\% & 0.188, ($6.31\times10^2$) \\
    & NAR4TSP & NAR & Sampling (1000) & 5.705 & 0.28\% & 0.047, ($\bm{1.88\times10^1}$) & 7.827 & 0.80\% & \textbf{0.113}, ($\bm{7.26\times10^1}$)\\
    \cmidrule{2-10}
     & AM  & AR & Sampling (2000) & 5.714 & 0.45\% & 0.140, ($1.39\times10^3$) & \textbf{7.934} & \textbf{2.18\%} & 0.399, ($3.98\times10^3$)  \\
     & GNARKD-AM & NAR  & Sampling (2000) & \textbf{5.706} & \textbf{0.30\%} & \textbf{0.040}, ($\bm{1.81\times10^1}$)  & 7.945 & 2.32\% & \textbf{0.067}, ($\bm{6.98\times10^1}$)  \\
    \cmidrule{2-10}
     &POMO & AR  & Greedy ($\times$8 augment) & \textbf{5.689} & \textbf{0.02\%} & 0.063, ($3.21\times10^1$) & \textbf{7.771} & \textbf{0.09\%} & 0.104, ($1.17\times10^2$) \\
    &GNARKD-POMO  & NAR & Sampling (1000) & 5.691  & 0.05\%  & \textbf{0.032}, ($\bm{9.70\times10^0}$) &7.790   & 0.33\%  & \textbf{0.073}, ($\bm{3.58\times10^1}$) \\
    \cmidrule{2-10}
     &TM & AR  & Sampling (2000) & 5.690 & 0.02\% & 0.422, ($2.99\times10^3$) & 7.799 & 0.44\% & 1.36, ($1.22\times10^4$) \\
     &GNARKD-TM  & NAR & Sampling (2000) & \textbf{5.689} & \textbf{0.01\%} & \textbf{0.043}, ($\bm{1.88\times10^1}$) & \textbf{7.797} & \textbf{0.42\%} & \textbf{0.078}, ($\bm{7.17\times10^1}$) \\
    \midrule
    \multirow{6}{*}{\rotatebox{90}{CVRP}}   & LKH3 & exact solver & - & \textbf{10.360} & \textbf{0.00\%} &34.12, (2.81$\times 10^4$)  & \textbf{15.646} & \textbf{0.00\%} &65.85, ($5.27\times 10^4$)  \\
    & MDAM & AR & Sampling (30) & 10.498 & 1.34\% & 19.94, ($5.98\times 10^2$) & 16.033 & 2.47\% & 38.24, ($1.93\times 10^3$) \\
    \cmidrule{2-10}
    &POMO & AR  & Greedy ($\times$8 augment) & \textbf{10.699} & \textbf{3.28\%} & 0.095, ($3.85\times10^1$)  & \textbf{15.750} & \textbf{0.69\%} & 0.172, ($1.40\times10^2$) \\
     &GNARKD-POMO  & NAR & Sampling (1000) & 10.737  & 3.64\%  & \textbf{0.081}, ($\bm{1.62\times10^1}$) &16.240   & 3.80\%  & \textbf{0.142}, ($\bm{5.12\times10^1}$) \\
     \cmidrule{2-10}
     &AM & AR  & Sampling (2000) & \textbf{10.602} & \textbf{2.34\%} & 0.177, ($1.76\times10^3$)  & \textbf{16.182} & \textbf{3.43\%} & 0.511, ($5.02\times10^3$) \\
     &GNARKD-AM  & NAR & Sampling (2000) & 10.754 & 3.81\% & \textbf{0.083}, ($\bm{3.13\times10^1}$)  & 16.821 & 7.51\% & \textbf{0.144}, ($\bm{9.54\times10^1}$)  \\
    \bottomrule
    \end{tabular}}%
      \caption{Experiment results of GNARKD student models and baseline models on solving VRPs. The optimality gap is computed w.r.t. the exact solvers Concorde~(for TSP) and LKH3~(for CVRP); $^*$For the computation time, we report the average time to solve one single instance before the comma and the total time to solve all 10,000 instances within the parentheses.}
  \label{table2}%
\end{table*}%

\subsection{Performance Analysis of GNARKD}
\noindent We apply GNARKD to AM, POMO and TM for performance evaluations, each time using a test dataset comprising 10,000 randomly generated VRP instances, following the same approach of \cite{Hudson2022}.

Table~\ref{table1} presents the performance gap between students and their respective teachers in the same inference mode. The performance evaluation metrics include solution quality, time required to solve a single instance and all 10,000 instances. Note that because neural construction models simultaneously handle multiple input instances, the time required to solve 10,000 instances is significantly shorter than the time required to solve an arbitrary instance for 10,000 times. It is worthy highlighting that the three teachers employed in this study have previously demonstrated as achieving the fastest inference speed \cite{Bi2022}. However, as shown in Table~\ref{table1}, the GNARKD students, especially GNARKD-AM and GNARKD-TM, exhibit faster inference speed than their respective teachers and even achieve higher solution quality. The overall performance of GNARKD is beyond expectation considering that the students are under full supervision of their respective teachers. One key reason why GNARKD obtains high-performance student models is we set a low temperature when distilling the teacher models, whereby the students correctly learn a more leptokurtic action probability distribution, leading to the selection of more confident actions (see Figure~\ref{fig_temp} at the end of this experiment section). Our experimental results showcase the efficacy of GNARKD in transforming an AR model into an NAR one, resulting in significantly improved inference speed while achieving on-par solution quality. Additionally, Figures~\ref{fig5a} and~\ref{fig5c} provide evidence of the successful knowledge transformation from the teacher to the student, manifested by the remarkable similarity between their action probability distributions. Specifically, in comparison to GCN, GNARKD-TM exhibits significantly greater confidence, with 96\% of its action probabilities exceeding the probability threshold of 0.75.

Table~\ref{table2} presents a comparison of the performance between the GNARKD students and other baselines, including the exact solvers such as Concorde \cite{Applegate2007} and LKH3 \cite{Helsgaun2017}, and SOTA neural construction models such as MDAM \cite{Xin2021}, GCN \cite{Joshi2019}, and NAR4TSP \cite{Xiao2023}. As shown in Table~\ref{table2}, despite being limited by the performance of their respective teachers, the GNARKD students achieve competitive results with a significant inference speed advantage comparing to other baseline models. For TSPs, GNARKD-TM even achieves SOTA results for TSP-50. Although the GNARKD students do not outperform the baselines in terms of solution quality for CVRPs, the difference is negligible. Furthermore, the GNARKD students constitute a pioneer successful attempt to solve CVRPs using NAR approaches, providing a valuable baseline for future studies. Our results demonstrate that GNARKD students can be readily deployed in real-world scenarios demanding near-optimal VRP solutions with immediate availability, e.g., warehouses operated by autonomous mobile robots.

\begin{figure}[!t]
\centering
\subfigure[\tiny{TSP instances of varying size }]{\includegraphics[width=0.495\columnwidth]{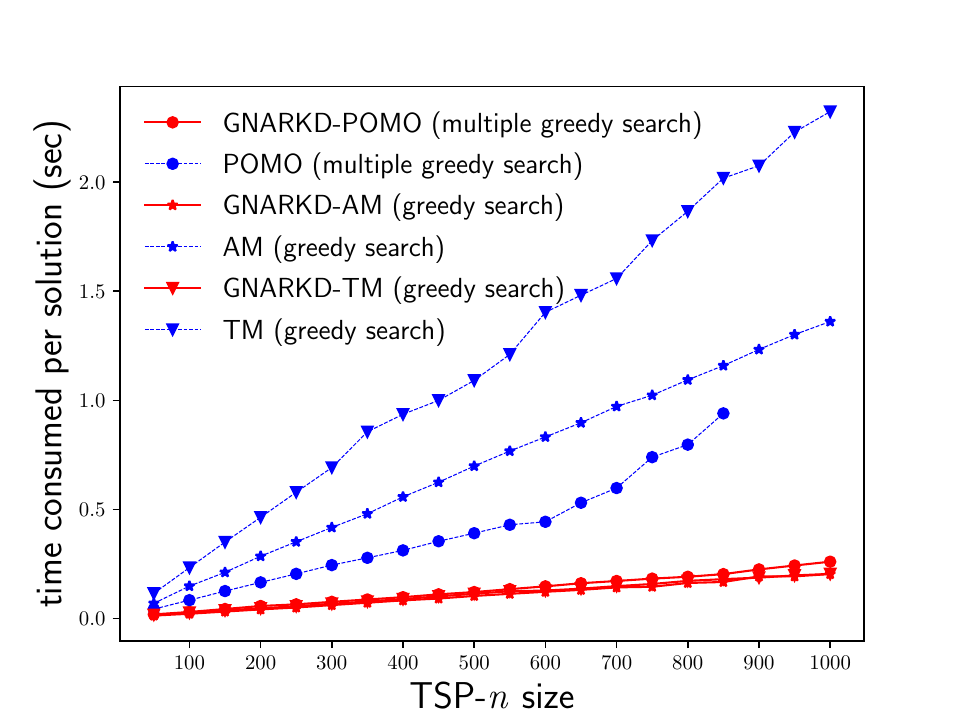}}
\subfigure[\tiny{CVRP instances of varying size}]{\includegraphics[width=0.495\columnwidth]{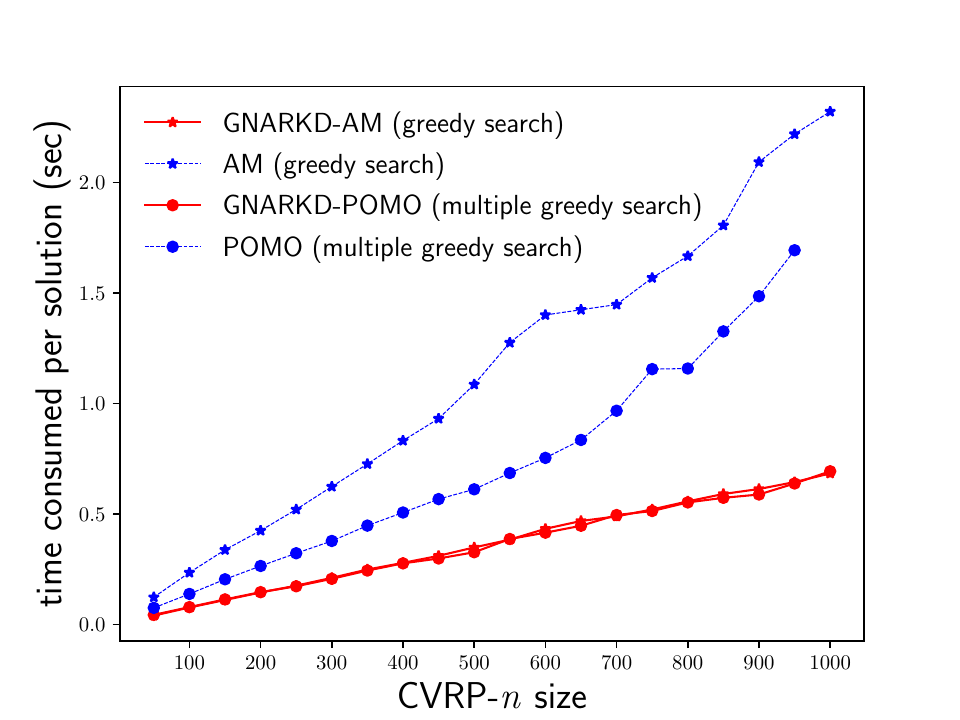}}
\subfigure[\tiny{TSP, beam search of varying width}]{\includegraphics[width=0.495\columnwidth]{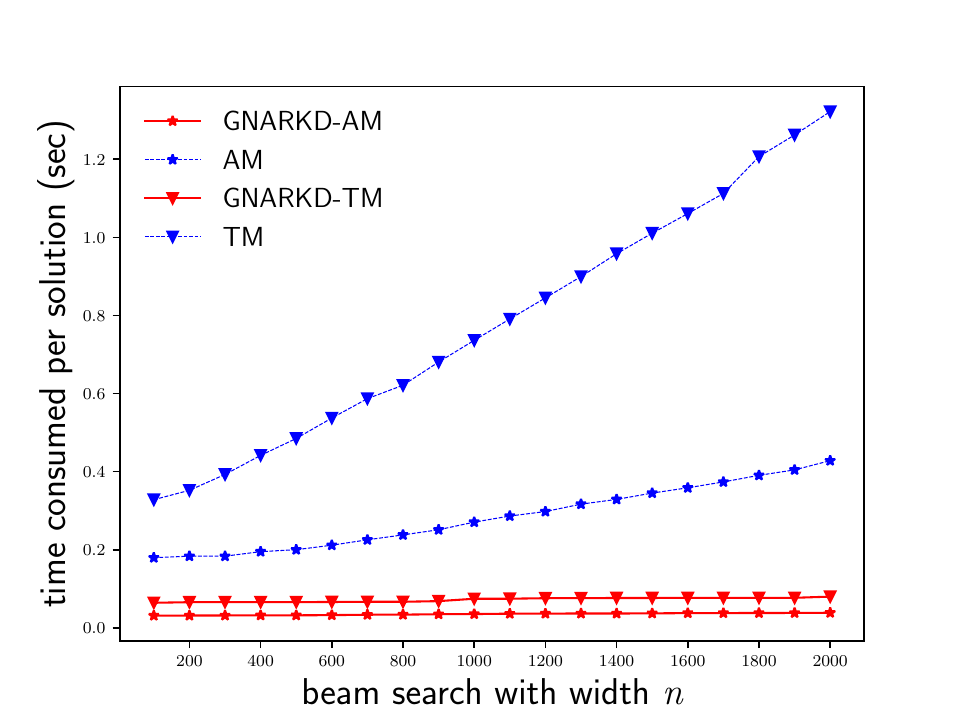}\label{beam1}}
\subfigure[\tiny{CVRP, beam search of varying width}]{\includegraphics[width=0.495\columnwidth]{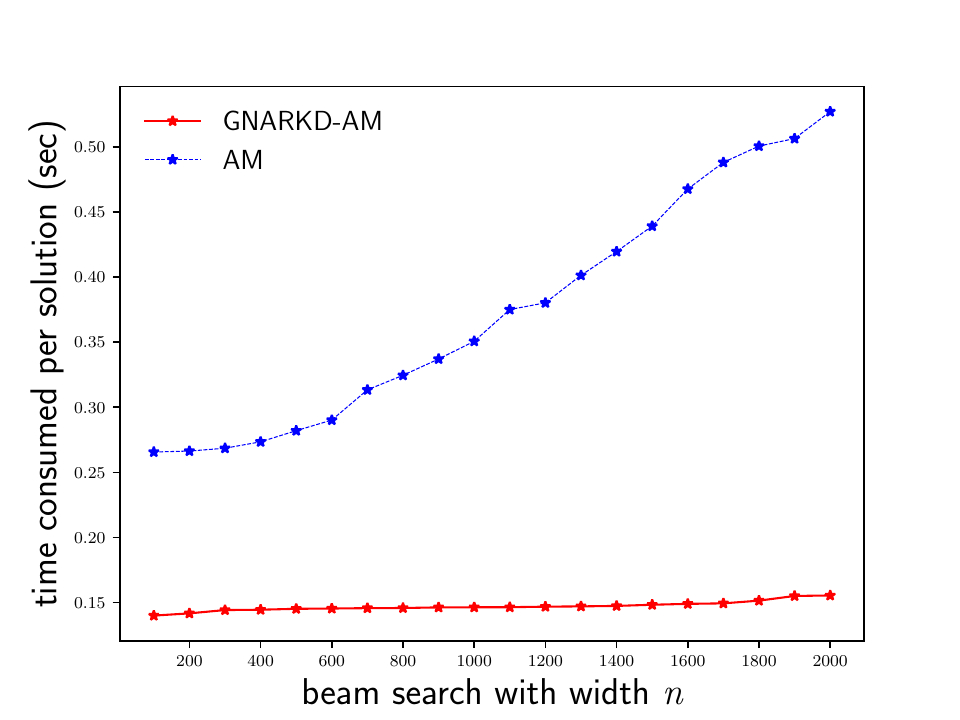}\label{beam2}}

\caption{Comparison on the execution time between GNARKD students and their respective teachers. Because POMO uses the multiple greedy rollouts instead of beam search, we only report the time taken by GNARKD-POMO using the same inference method. Moreover, constrained by GPU memory, POMO is unable to solve TSP instances larger than 850 and CVRP instances larger than 1,000.}\label{fig_size}
\end{figure}

\subsection{Inference Speed Analysis of GNARKD}
\noindent To highlight the remarkably fast inference speed of the GNARKD students, we conduct two more sets of experiments. First, we compare the solving time of the student and teacher models for VRP instances of varying sizes using the greedy mode. Secondly, we compare the solving time of the student and teacher models for a predetermined VRP instance using beam search with varying widths. We repeat each experiment for 100 times and report the average results in Figure~\ref{fig_size}. As shown, the GNARKD students have a significantly faster inference speed across all experiments when compared to their respective teachers. The speed further elevates with the increase of an instance size or beam width. These results showcase the speed advantages of employing the student model over the teacher model, again indicating its great potential for deployment in real-world scenarios.

\begin{table}[!t]
      \centering
    \resizebox{1\columnwidth}{!}{
    \begin{tabular}{cccc}
    \toprule
        \multicolumn{2}{c}{\multirow{2}{*}{\textbf{Algorithms}}} & \tabincell{c}{Average \\optimality gap} $\downarrow$ & \tabincell{c}{Average \\time (Sec)} $\downarrow$ \\
        \midrule
        \multicolumn{2}{c}{Concorde} & 0.00\% & 0.178 \\
        \midrule
        \multirow{2}{*}{AM} & Greedy  & 7.96\% & 0.153 \\
        ~ & Sampling~(B=2000) & 6.93\% & 0.456\\
        \multirow{2}{*}{GNARKD-AM} & Greedy & 10.98\% & 0.024 \\
        ~ & Sampling~(B=2000) & 5.88\% & 0.075 \\
        \midrule
        \multirow{2}{*}{TM} & Greedy & 8.60\% & 0.281 \\
        ~ & Sampling~(B=2000) & 5.17\% & 1.565 \\
        \multirow{2}{*}{GNARKD-TM} & Greedy & 11.92\% & 0.027 \\
        ~ & Sampling~(B=2000) & 3.89\% & 0.076 \\
        \midrule
        \multirow{2}{*}{POMO} & Greedy~(no augment) & 3.41\% & 0.090  \\
        ~ & Greedy~($\times8$ augment) & 2.53\% & 0.103 \\
        \multirow{2}{*}{GNARKD-POMO} & Greedy~(no augment) & 6.41\% & 0.036 \\
        ~ & Greedy~($\times8$ augment) & 4.49\% & 0.037 \\
        \bottomrule
        \end{tabular}}
        \caption{Performance comparison on 20 TSPLIB instances}\label{TSPLIB}
\end{table}

\subsection{Further Performance Analysis of GNARKD}\label{necessity}

\noindent To further evaluate the performance of GNARKD in real-world scenarios, we randomly choose 20 real-world TSP instances (node sizes ranging from 51 to 159) from TSPLIB as the test cases. In addition to reporting the solution length, we also report the execution time for a comprehensive comparison. Furthermore, we use the exact solver Concorde that produces optimal results as the baseline.

As shown in Table~\ref{TSPLIB}, both AM and TM do not exhibit any advantage in terms of solution quality and execution time compared to Concorde. In this case, the use of Concorde is more preferable than employing these two models. Conversely, the GNARKD students exhibit significantly shorter inference time compared to Concorde. This finding indicates that GNARKD improves the effectiveness of these two models, emphasizing the significance of our method.

Although POMO exhibits remarkable performance, its solution is generated by multiple greedy rollouts, demanding greater computational resources than the single greedy rollouts used in conventional AR models. In contrast, GNARKD-POMO generates solutions exclusively within the output matrix $A$, resulting in substantial saving in computational resources. To assess this disparity, we examine the maximum node size that POMO and GNARKD-POMO can handle. Our findings, constrained by an 8GB GPU memory limitation, indicate that POMO can solve TSP instances with a maximum of 1,000 nodes when employing multiple greedy rollouts, while GNARKD-POMO can solve TSP instances with up to 9,000 nodes. When using multiple greedy rollouts with $\times8$ augments, POMO's capacity is restricted to TSP instances of up to 500 nodes, whereas GNARKD-POMO can solve TSP instances with up to 3,100 nodes. These findings substantiate that our student model is more suitable for practical applications with limited resources.

\subsection{Ablation Study of GNARKD}\label{distill}

\noindent To assess the effectiveness of the distillation temperature and the guided KD, we conduct two ablation studies as follows.

\begin{table}[!t]
  \centering

    \resizebox{0.8\columnwidth}{!}{
  \begin{tabular}{cccc}
   \toprule
      & SL & RL & Guided KD \\
    \midrule
    GNARKD-AM & 9.30\% & 2.42\% & \textbf{0.36\%} \\
    \midrule
    GNARKD-TM & 7.69\% & 1.83\% & \textbf{0.01\%} \\
    \midrule
    GNARKD-POMO & 7.35\% & 1.99\% & \textbf{0.05\%} \\
    \bottomrule
\end{tabular}}
  \caption{Optimality gap of student models trained with different methods using beam (1000) search for TSP-50s}\label{diff_table}
\end{table}

\begin{figure*}[!t]
\centering
\subfigure[\tiny GNARKD-AM, TSP]{\includegraphics[width=0.68\columnwidth]{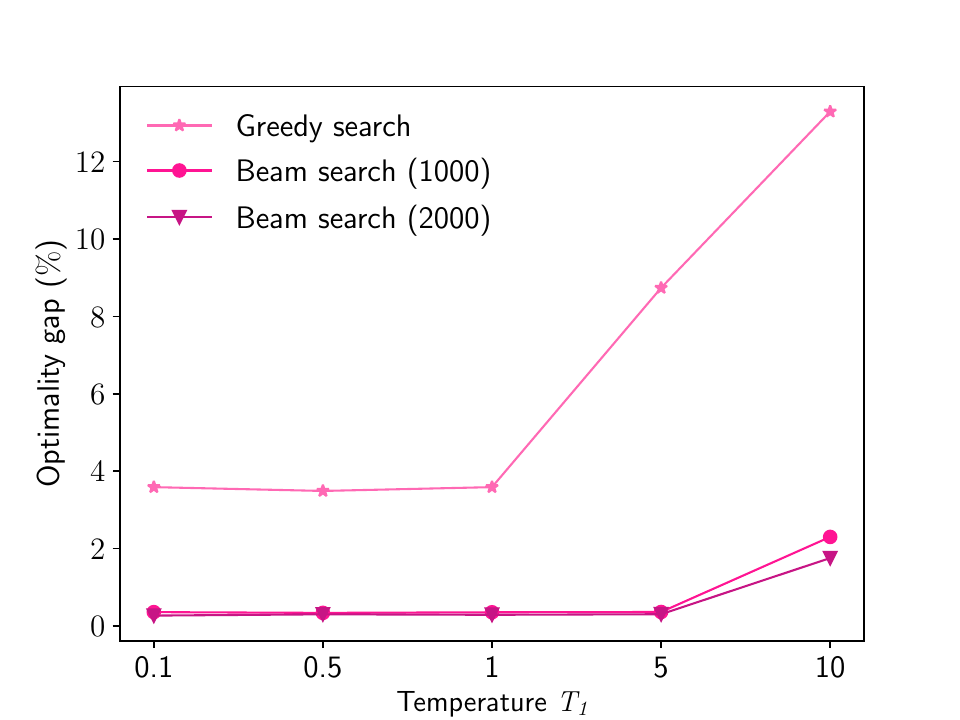}}
\subfigure[\tiny GNARKD-TM, TSP]{\includegraphics[width=0.68\columnwidth]{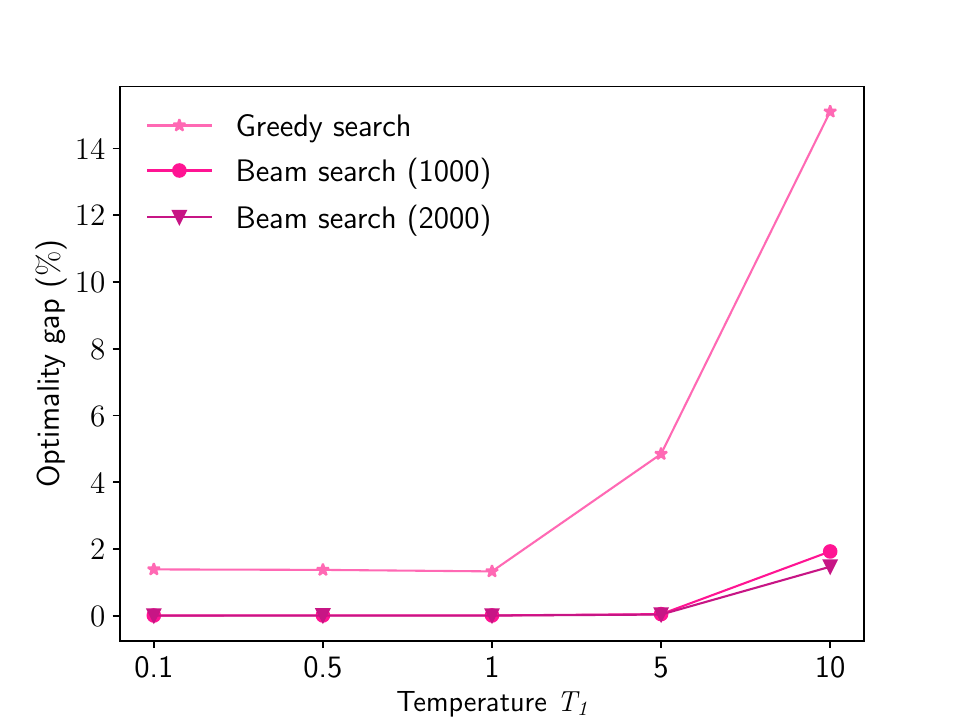}}
\subfigure[\tiny GNARKD-POMO, TSP]{\includegraphics[width=0.68\columnwidth]{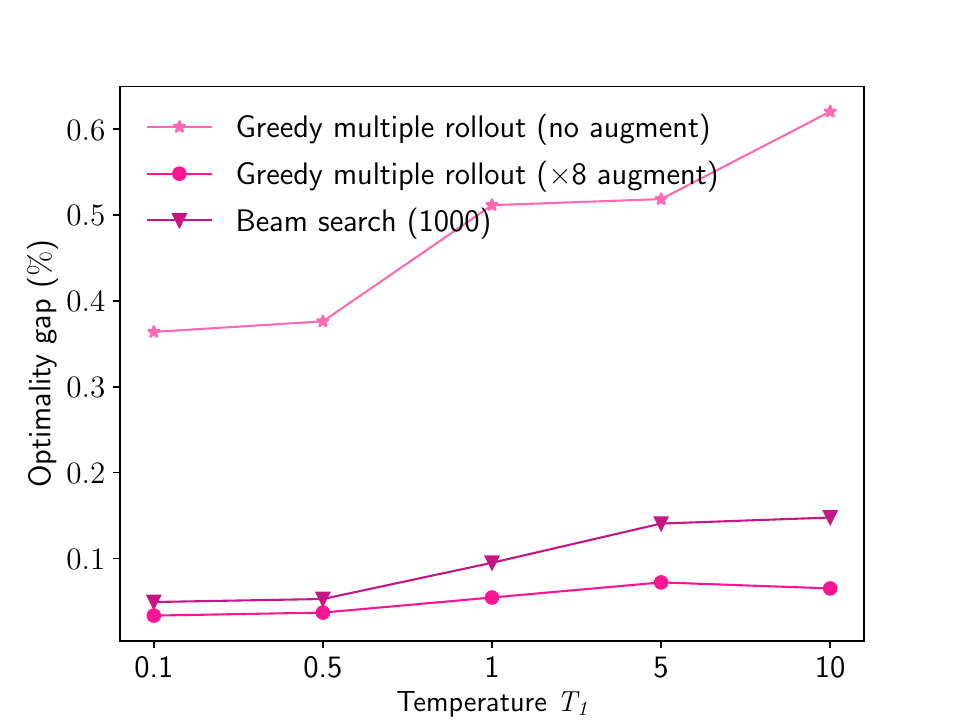}}
\subfigure[\tiny GNARKD-AM, CVRP]{\includegraphics[width=0.68\columnwidth]{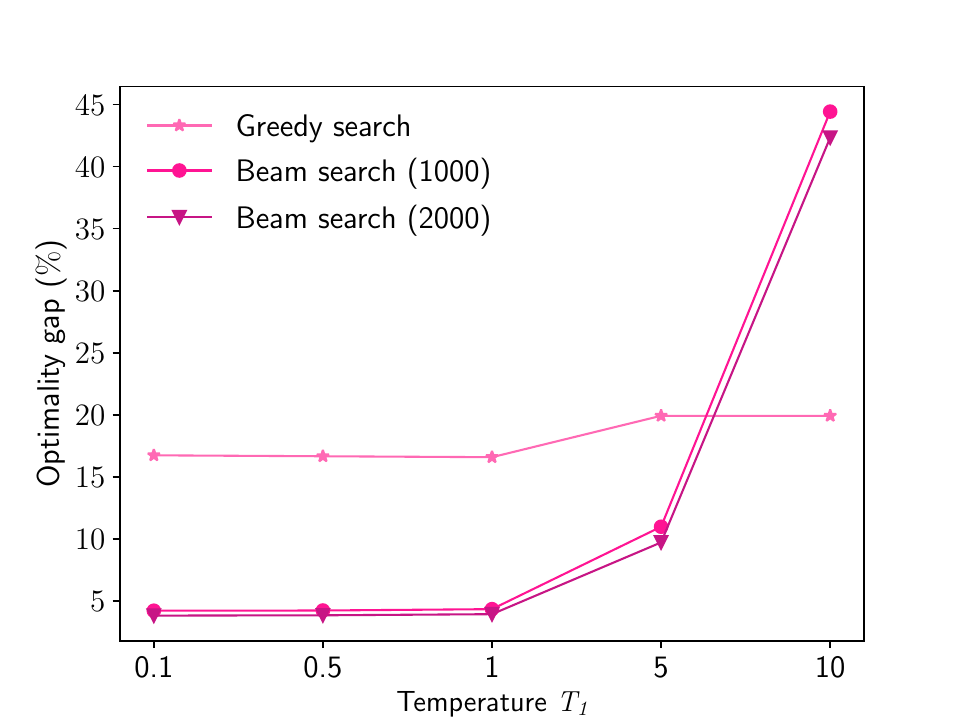}}
\subfigure[\tiny GNARKD-POMO, CVRP]{\includegraphics[width=0.68\columnwidth]{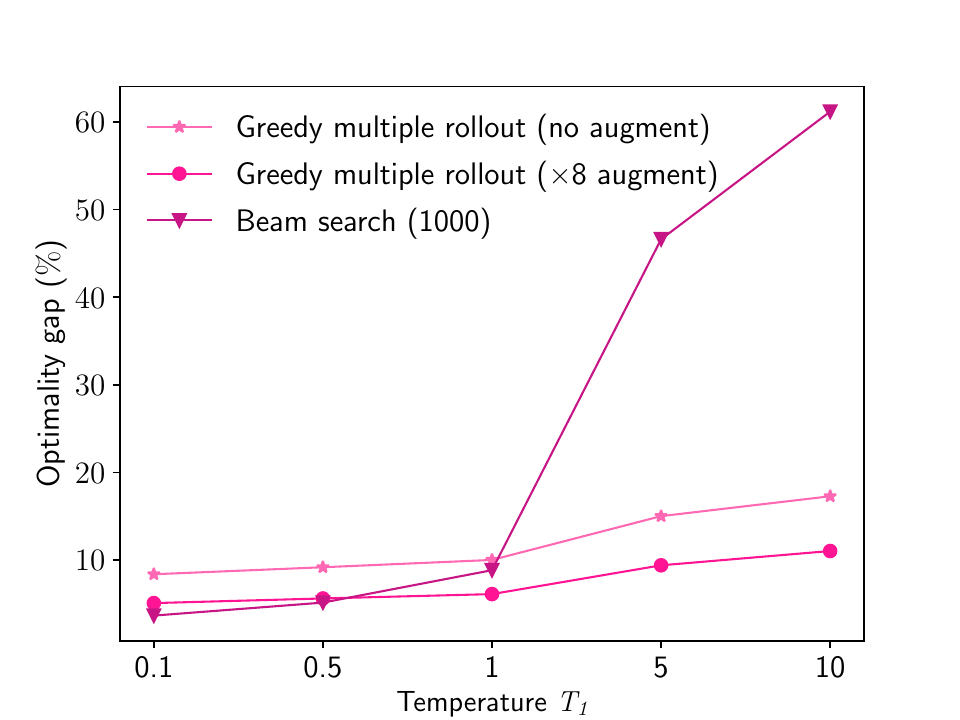}}
\caption{Performance of GNARKD students under different distillation temperatures.}\label{fig_temp}
\end{figure*}

\begin{figure*}[!t]
\centering
\subfigure[$T_1=0.1$]{\includegraphics[width=0.68\columnwidth]{Figure7a.pdf}}
\subfigure[$T_1=0.5$]{\includegraphics[width=0.68\columnwidth]{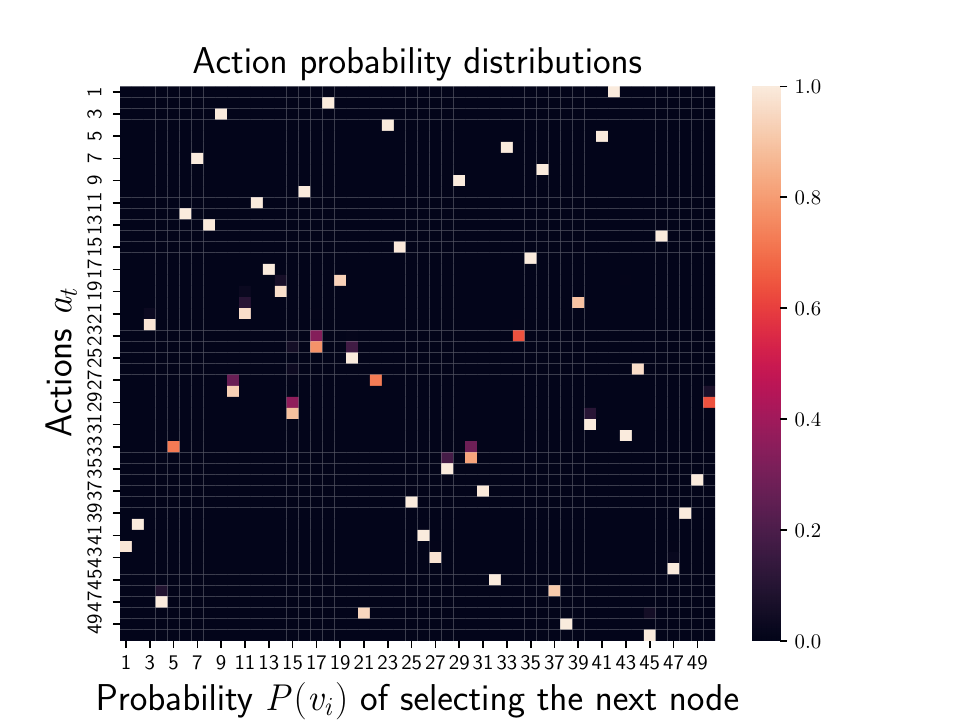}}
\subfigure[$T_1=1$]{\includegraphics[width=0.68\columnwidth]{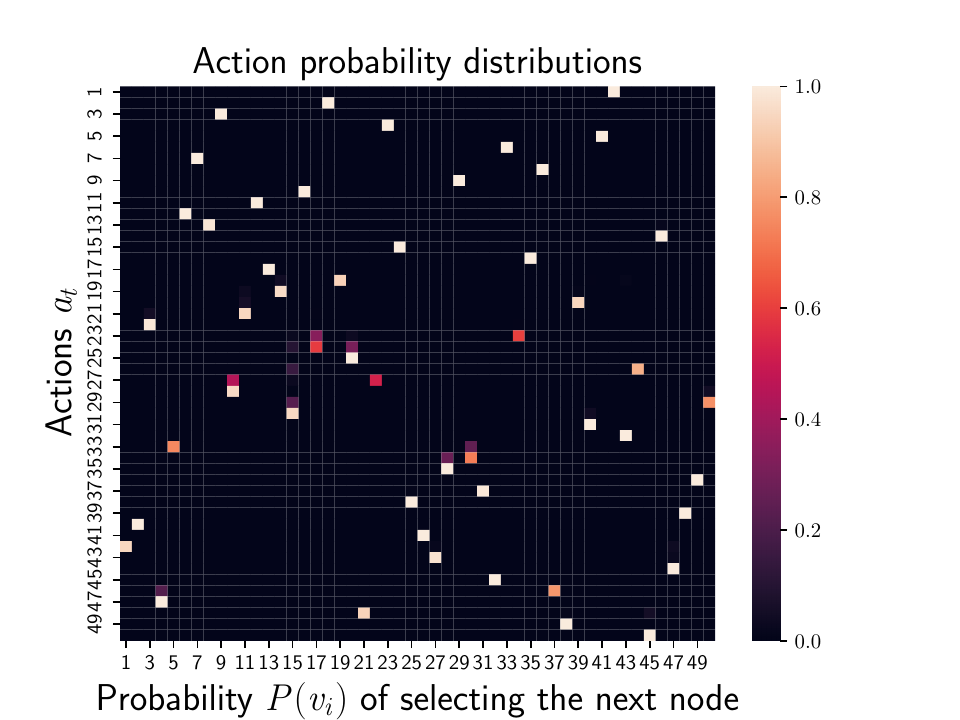}}
\subfigure[$T_1=5$]{\includegraphics[width=0.68\columnwidth]{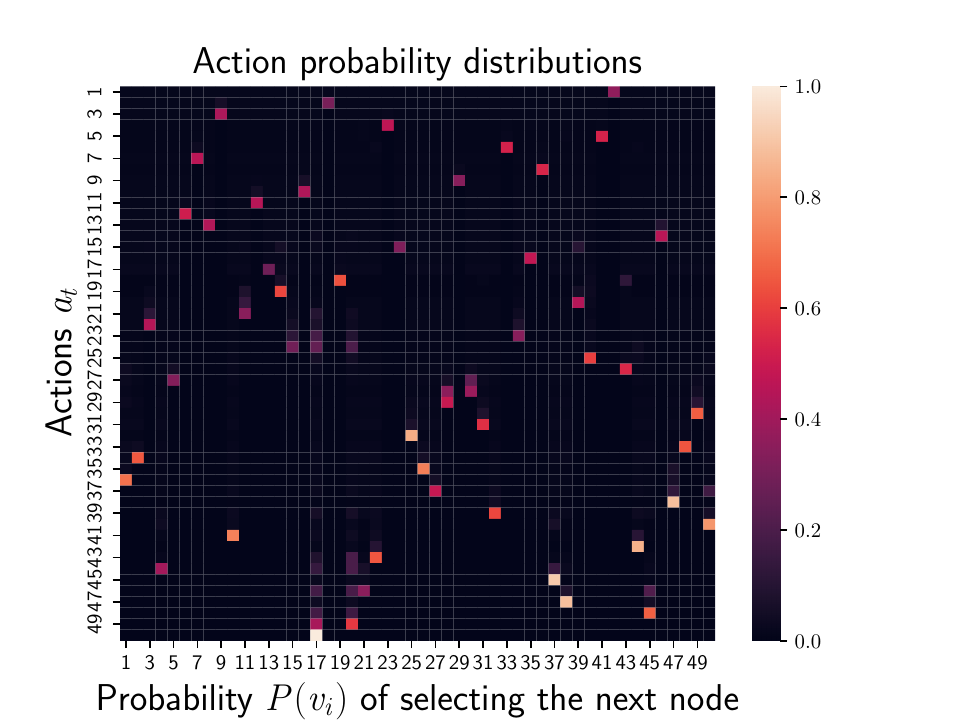}}
\subfigure[$T_1=10$]{\includegraphics[width=0.68\columnwidth]{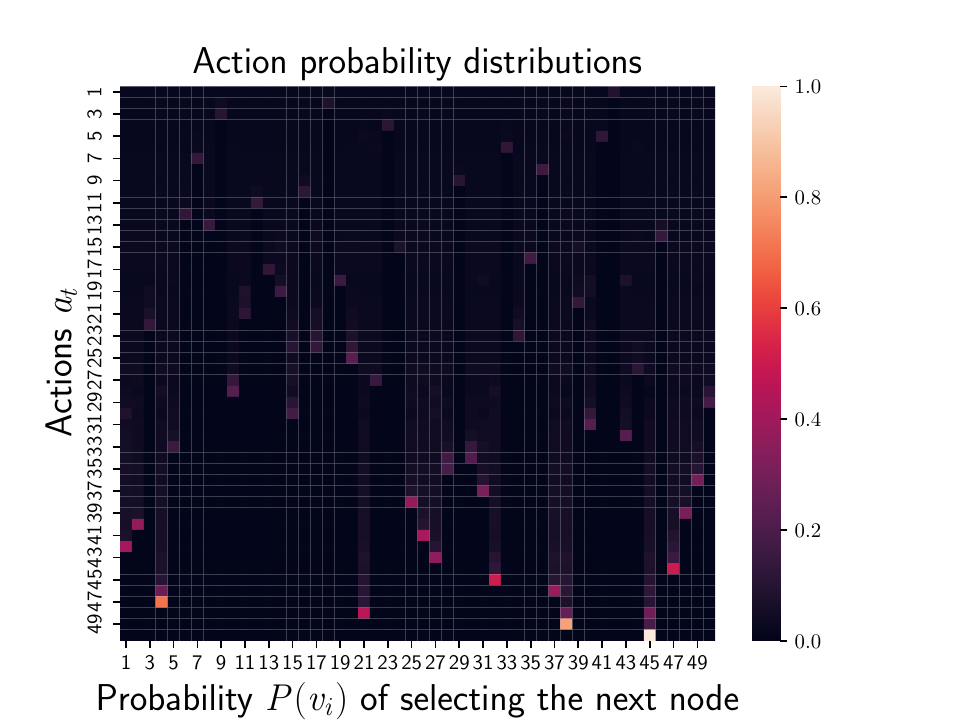}}
\caption{Action probability distribution of GNARKD-TM trained with different temperature $T_1\in \{0.1, 0.5, 1, 5, 10\}$ when solving a randomly generated TSP-50 instance.}\label{Vis_tem}
\end{figure*}

\subsubsection{Sensitivity on Distillation Temperature.} We conduct ablation studies on VRP-50 to analyze whether the KD temperature significantly affects the performance of the students. Specifically, we use different values of $T_1\in\{0.1, 0.5, 1, 5, 10\}$ during training and the results (see Figure~\ref{fig_temp}) indicate that for low temperatures $T_1\in\{0.1, 0.5, 1\}$, the performance difference among the students trained at different temperatures are insignificant, primarily due to the leptokurtic probability distribution of actions observed in the teachers. Conversely, using a higher temperature $T_1\in \{5, 10\}$ for distillation leads to a notable decline in the performance of the students. Furthermore, we visualize the action probability distribution of GNARKD-TM trained at different temperatures when solving a randomly generated TSP-50 instance~(see Figure~\ref{Vis_tem}).  Our statistical analysis indicates that the lower bound probabilities of actions taken by GNARKD-TM trained at a low temperature $T_1\in \{0.1, 0.5, 1\}$ are significantly higher ($\{0.65, 0.64, 0.53\}$) compared to GNARKD-TM trained at a high temperature $T_1\in \{5, 10\}$, which exhibit lower bound probabilities of $\{0.28, 0.07\}$. Furthermore, when using a probability threshold of 0.75 to identify confident actions, GNARKD-TM trained at a varying temperature $T_1\in \{0.1, 0.5, 1, 5, 10\}$ shows proportions of confident actions at $96\%$, $92\% $, $92\%$, $14\%$ and $4\%$, respectively. These results substantiate the efficacy of GNARKD in training students to learn an accurate leptokurtic action probability distribution, resulting in enhanced performance.

\subsubsection{Effectiveness of Different Training Methods.} We compare the performance of the student models trained with different methods for TSP-50 to evaluate the effectiveness of our guided KD. Specifically, we utilize the SL and RL training methods outlined in prior studies \cite{Joshi2019} and \cite{Xiao2023}, respectively, for training our student models. Both training methods are conducted within the same training environment utilized for the proposed guided KD. The results presented in Table~\ref{diff_table} demonstrate that the models trained using SL yield suboptimal performance, whereas those trained using RL show significant performance improvements. However, the models trained with guided KD outperforms them both. These results demonstrate the high efficacy of training students using our guided KD training method.

\section{Conclusion and Future Work}
This paper proposes GNARKD, a technique for transforming the knowledge of Transformer-based AR models into NAR models, which effectively utilize the high parallelism of the Transformer during inference. We apply GNARKD to three prominent AR models and present comprehensive experimental results with comparisons. The results show that GNARKD significantly improves the inference speed while maintaining competitive solution quality.

The current version of GNARKD may not perform well with complex constraints, and the student's performance is inevitably limited by its respective teacher. Going forward, we plan to 1)~model the constraints on NAR models directly on the decoder instead of post-processing the model output and 2)~use multiple teachers to train students for further performance gain.

\section{Acknowledgments}
This work is supported by the National Key Research and Development Program of China (2021YFF1201200), the Jilin Provincial Department of Science and Technology Project (20230201083GX, 20220201145GX, and 20230201065GX), the National Natural Science Foundation of China (62072212, 61972174, 61972175, and 12205114), the Guangdong Universities’ Innovation Team Project (2021KCXTD015), the National Scholarship Funds, Key Disciplines (2021ZDJS138) Projects, the Nanyang Associate Professorship, and the National Research Foundation Fellowship (NRF-NRFF13-2021-0006), Singapore. Any opinions, findings, conclusions, or recommendations expressed in this material belong to the authors and do not reflect the views of the funding agencies.

\bibliography{MyBib1}

\end{document}